\documentclass[11pt, letterpaper]{article}

\usepackage{fullpage}
\usepackage{graphicx}
\usepackage{amsmath, amssymb}
\usepackage{amsthm}
\usepackage{dirtytalk}
\usepackage{xfrac}
\usepackage{xcolor}

\usepackage[round]{natbib}

\usepackage[pagebackref=true]{hyperref}

\let\cite\citep
\definecolor{darkscarlet}{rgb}{0.34, 0.01, 0.1}
\definecolor{yaleblue}{rgb}{0.06, 0.3, 0.57}
\definecolor{darkpowderblue}{rgb}{0.0, 0.2, 0.6}

\hypersetup{
    colorlinks=true,
    citecolor=darkscarlet,
    linkcolor=darkpowderblue,
    filecolor=magenta,      
    urlcolor=yaleblue,
    menucolor=gray
}

\DeclareMathOperator{\oalpha}{\overline{\alpha}}
\newcommand{\gf}{\nabla f_\xi}

\usepackage[hyperpageref]{backref}
\renewcommand*{\backref}[1]{}
\renewcommand*{\backrefalt}[4]{%
   \ifcase #1 %
     \footnotesize{(Not cited.)}%
   \or
     \footnotesize{(Cited on page~#2)}%
   \else
     \footnotesize{(Cited on page~#2)}%
\fi }

\definecolor{midnightblue}{HTML}{0059b3}
\usepackage{xspace}
\newcommand{\algname}[1]{{\color{midnightblue!70!black}\small\sf#1}\xspace}
\newcommand{\alg}{\algname{QC-SGD}}
\newcommand{\sgd}{\algname{SGD}}

\usepackage{mathtools, amsthm, amsmath, amssymb}
\mathtoolsset{showmanualtags}
\usepackage{xcolor}
\usepackage{tabularx}
\usepackage{bbm}
\usepackage{makecell}

\usepackage[utf8]{inputenc} 
\usepackage[T1]{fontenc}    
\usepackage{url}            
\usepackage{booktabs}       
\usepackage{amsfonts}       
\usepackage{nicefrac}       
\usepackage{microtype}      
\usepackage{xcolor}         

\usepackage{xspace}
\usepackage{xfrac}
\usepackage{verbatim}

\usepackage{algorithm}
\usepackage{algpseudocode}
\usepackage{todonotes}
\usepackage{subcaption}

\usepackage{nicefrac}

\allowdisplaybreaks[1]

\newtheorem{theorem}{Theorem}
\newtheorem{lemma}{Lemma}
\newtheorem{corollary}{Corollary}

\newtheorem{assumption}{Assumption}
\newtheorem{example}{Example}

\newcommand{\Exp}[1]{{\color{black}  \mathbb{E}}\left[#1\right]}

\newcommand{\Rd}{\mathbb{R}^d}
\newcommand{\R}{\ensuremath{\mathbb{R}}}

\def\<#1,#2>{\langle #1,#2\rangle}

\newcommand{\inner}[2]{\left\langle #1 ,  #2 \right\rangle}

\definecolor{burgundy}{rgb}{0.5, 0.0, 0.13}
\definecolor{darkscarlet}{rgb}{0.34, 0.01, 0.1}

\newcommand{\norm}[1]{\left\|#1\right\|}

\newcommand\br[1]{\left( #1 \right)}

\newcommand\cbr[1]{\left\{#1\right\}}

\newcommand{\sqn}[1]{\norm{#1}^2}

\newcommand{\EE}{\mathbb{E}}

\newcommand{\II}{\mathbb{I}}

\newcommand{\RR}{\mathbb{R}}

\newcommand{\cD}{\mathcal{D}}

\newcommand{\cO}{\mathcal{O}}

\newcommand{\cQ}{\mathcal{Q}}

\newcommand{\eqdef}{\coloneqq}

\usepackage{pifont}
\newcommand{\circledOne}{\text{\large\ding{172}} }
\newcommand{\circledTwo}{\text{\large\ding{173}} }
\newcommand{\circledThree}{\text{\large \ding{174}} }

\title{On the Convergence of DP-SGD with Adaptive Clipping}
\author{%
    Egor Shulgin
  \qquad
  Peter Richt{\'{a}}rik  \vspace{5pt} \\
  King Abdullah University of Science and Technology (KAUST) \\
  Thuwal, Saudi Arabia\\
}
\date{}

\begin{document}

\maketitle

\begin{abstract}%
Stochastic Gradient Descent (\sgd) with gradient clipping is a powerful technique for enabling differentially private optimization. Although prior works extensively investigated clipping with a constant threshold, private training remains highly sensitive to threshold selection, which can be expensive or even infeasible to tune. This sensitivity motivates the development of adaptive approaches, such as quantile clipping, which have demonstrated empirical success but lack a solid theoretical understanding. This paper provides the first comprehensive convergence analysis of \sgd with quantile clipping (\alg). We demonstrate that \alg suffers from a bias problem similar to constant-threshold clipped \sgd but show how this can be mitigated through a carefully designed quantile and step size schedule. Our analysis reveals crucial relationships between quantile selection, step size, and convergence behavior, providing practical guidelines for parameter selection. We extend these results to differentially private optimization, establishing the first theoretical guarantees for \algname{DP-QC-SGD}.
Our findings provide theoretical foundations for widely used adaptive clipping heuristic and highlight open avenues for future research.
\end{abstract}

\section{Introduction}
It is hard to imagine the success of modern Machine Learning without effective optimization, the cornerstone of which are Stochastic Gradient Descent (\sgd) type methods \cite{robbins1951stochastic, bottou2018optimization}. 
However, \sgd is not perfect, particularly in the context of Deep Learning. 
Efficient neural network training often requires modifications of \sgd to stabilize optimization. For example, exploding gradients issue \cite{pascanu2012understanding, pascanu2013difficulty} is often tackled by the use of \emph{gradient clipping} operator, which scales down the input vector's norm if it exceeds a certain threshold.
Moreover, gradient clipping is vital in privacy-preserving machine learning \cite{dwork2006calibrating, dwork2014algorithmic}. 
Rigorous differential privacy guarantees are usually established by relying on the Gaussian mechanism \cite{dwork2014algorithmic}, which requires bounded sensitivity to control the amount of noise added.
The most commonly used in practice Differentially Private SGD (\algname{DP-SGD}) method enforces such bound by clipping the per-example gradients \cite{abadi2016deep}.

Clipped \sgd was shown to be superior to vanilla \sgd for minimizing generalized smooth functions \cite{zhang2019gradient} and when stochastic gradient noise is heavy-tailed \cite{zhang2020adaptive}.
However, the effectiveness of gradient clipping hinges critically on the choice of the clipping threshold, denoted as $\tau$. This introduces an additional hyperparameter that requires careful tuning, a challenge especially pronounced in private optimization settings
where performance can be highly sensitive to this threshold \cite{kurakin2022toward, bu2024automatic}. Furthermore, each training run incurs an additional privacy loss, making extensive hyperparameter search prohibitively expensive from a privacy perspective \cite{papernot2021hyperparameter}.

\begin{figure*}[!hb]
\centering
    \includegraphics[width=0.31\textwidth, trim=0 75 0 0, clip]{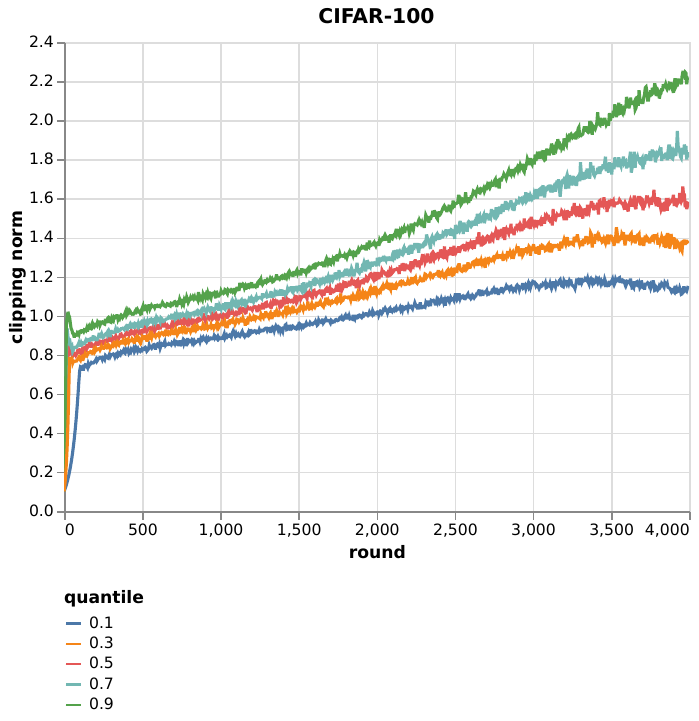}
\hfill
    \includegraphics[width=0.31\textwidth, trim=0 75 0 0, clip]{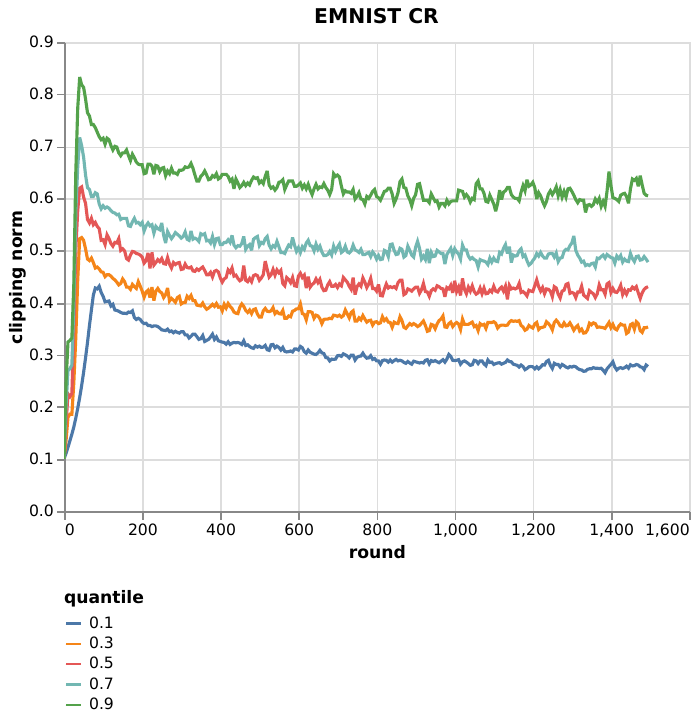}
\hfill
    \includegraphics[width=0.31\textwidth, trim=0 75 0 0, clip]{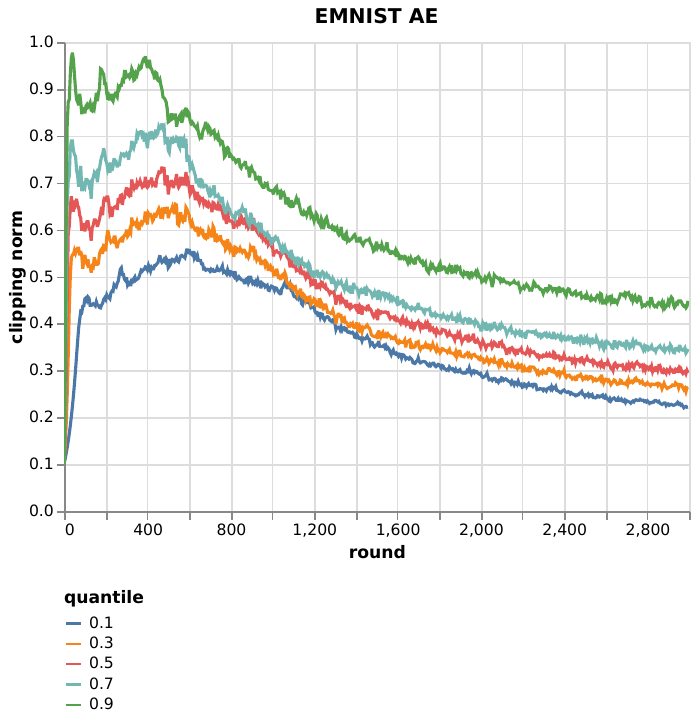}
    \\ 
\vspace{1ex}
    \includegraphics[width=0.31\textwidth, trim=0 75 0 0, clip]{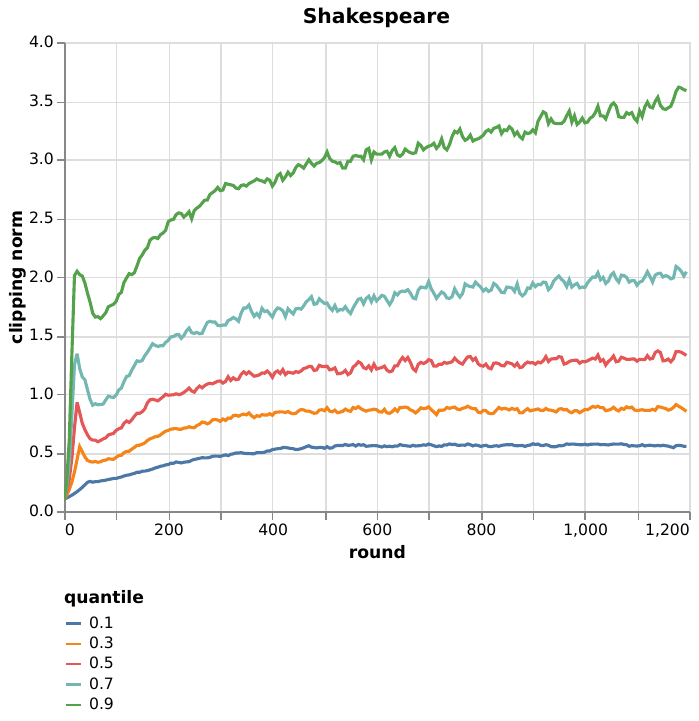}
\hfill
    \centering
    \includegraphics[width=0.31\textwidth, trim=0 75 0 0, clip]{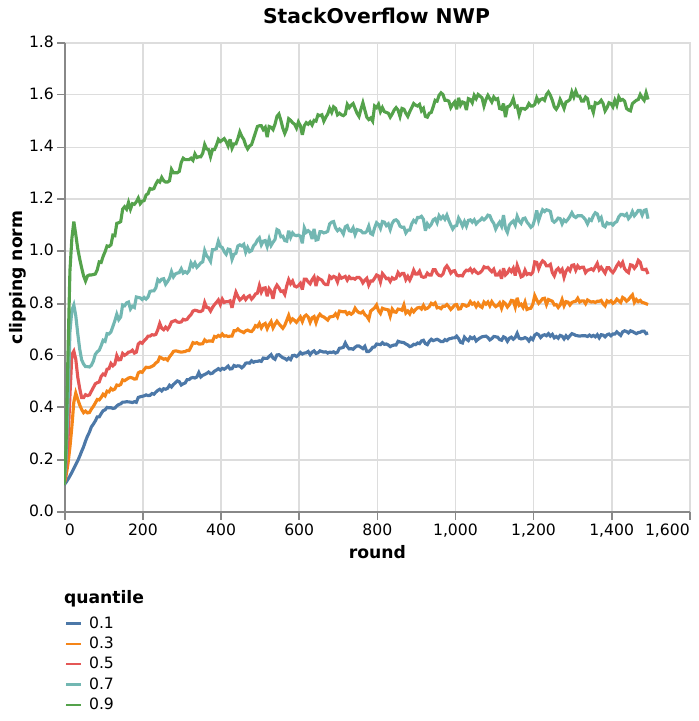}
\hfill
    \centering
    \includegraphics[width=0.31\textwidth, trim=0 75 0 0, clip]{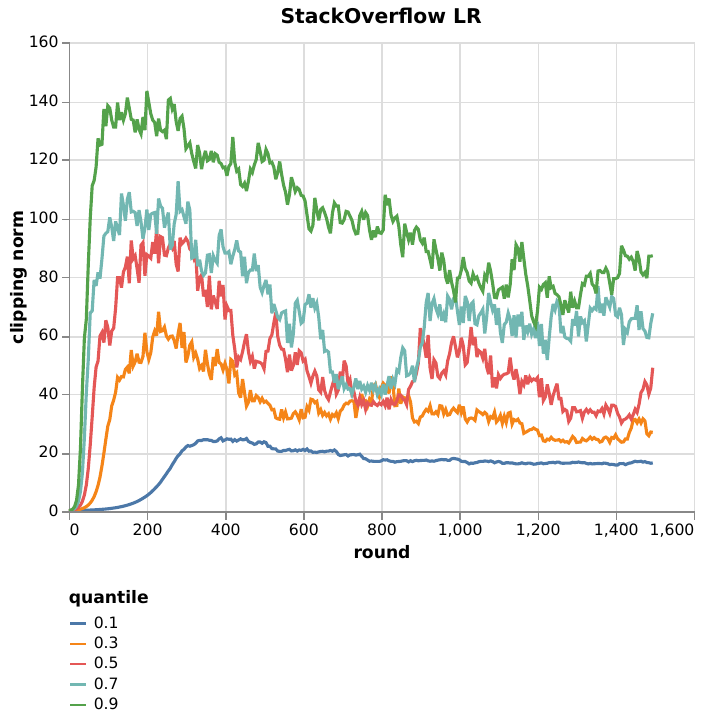}
\caption{Evolution of the adaptive clipping norm at five different quantiles (0.1, 0.3, 0.5, 0.7, 0.9) on six federated learning problems without Differential Privacy noise. Note that each task has a unique shape (e.g., increasing and decreasing) to its update norm evolution, which further motivates an adaptive approach. The figure is taken from the paper by \citet{andrew2021differentially}.}
    \label{fig:quantile_clip}
\end{figure*}

\paragraph{Adaptive clipping.} The problem described above has been addressed by \citet{andrew2021differentially} in the setting of private Federated Learning \cite{konecny2017federated, mcmahan2017communication, mcmahan2018learning, kairouz2021advances}. Specifically, \citet{andrew2021differentially} proposed to adaptively select the clipping threshold based on the distribution of gradient norms (or updates) of the participating clients. Their method (named \say{adaptive quantile clipping}) efficiently estimates a quantile and applies it as the clipping threshold. Crucially, privacy analysis by \citet{andrew2021differentially} revealed that this adaptive approach incurs only negligible additional privacy loss. 
Extensive empirical evaluations have shown that quantile clipping is competitive with and often outperforms carefully tuned constant clipping baselines \cite{andrew2021differentially}. This adaptive strategy offers the additional benefit of adjusting to the evolving gradient distribution throughout the federated optimization process. This adaptability is particularly valuable given the significant variability observed across different machine learning tasks and datasets (see Figure \ref{fig:quantile_clip}), suggesting that a uniform clipping schedule may be suboptimal.
The success of the adaptive clipping technique has led to its widespread adoption for multiple applications \cite{song2022flair, xu2023federated}, even beyond privacy-constrained settings \cite{charles2021large}, and implementation in Federated Learning libraries \cite{TensorFlow_Federated2018, granqvist2024pfl}.

However, despite its practical success and adoption, the theoretical properties of adaptive quantile clipping remain largely unexplored.
This research gap motivates our current study. We aim to provide a comprehensive optimization analysis of \sgd with quantile clipping (\alg in short).
In particular, by building upon recent work \cite{merad2024robust}, we demonstrate that \sgd with quantile clipping suffers from a bias problem, preventing convergence, similar to that observed in constant clipped \sgd. We design a quantile and step size schedule that effectively eliminates the identified bias problem. Our analysis reveals the crucial relationship between the chosen quantile value and the step size in \alg, providing insights into how this interplay affects convergence. Finally, we extend the results to differentially private optimization.

\section{Problem and Assumptions}

We consider a classical stochastic optimization problem
\begin{equation} \label{eq:problem} 
	\min_{x \in \RR^d} \Bigl[ f(x) \eqdef \EE_{\xi\sim \cD}[f_{\xi}(x)]\Bigr],
\end{equation}
where $f_\xi: \RR^d \to \RR$ is a loss of machine learning model parametrized by vector $x \in \RR^d$ on data point $\xi$. Thus, $f$ is excess loss over the distribution of data $\xi \sim \cD$. 
We rely on standard assumptions for non-convex stochastic optimization to enable optimization analysis.
\begin{assumption}[Stochastic gradient] \label{ass:b_variance}
Stochastic gradient estimator is unbiased $\Exp{\nabla f_{\xi}(x)} = \nabla f(x)$ and has \textbf{bounded $q$-th moment} for $q \in (1, 2]$ 
\begin{equation} \label{eq:bound_var}
    \br{\EE_{\xi \sim \cD} \left[ \left\| \nabla f_{\xi}(x) - \nabla f(x) \right\|^q \right]}^{1/q} \leq \sigma_q, \qquad \forall x\in \R^d.
\end{equation}
\end{assumption} 
Condition \eqref{eq:bound_var} is usually referred as heavy tailed noise \cite{zhang2020adaptive} for $q \in (1, 2)$. For $q=2$, it recovers classical bounded variance assumption \cite{nemirovski2009robust, ghadimi2012optimal}.

\begin{assumption}[Function] \label{ass:l_smooth}
    The function \( f \) is differentiable and \( L \)-\textbf{smooth}, meaning there exists \( L > 0 \) such that 
    \begin{equation}
        f(x+u) \leq f(x) + \langle \nabla f(x), u \rangle + \frac{L}{2} \|u\|^2, \qquad \forall x,u \in \R^d.
    \end{equation}
    Additionally, \( f \) is lower-bounded by \( f^{\inf} \in \mathbb{R} \).
\end{assumption}

\section{Stochastic Gradient Descent with Quantile Clipping}

We consider the following Stochastic Gradient Descent type method with step size $\gamma_t > 0$
\begin{equation} \label{eq:sgd_qc}
    x^{t+1} = x^t - \gamma_t g^t,
\end{equation}
where $g^t = g(x^t)$ has the form of a clipped stochastic gradient estimator $g^t = \alpha_{\xi^t}(x^t) \nabla f_{\xi^t}(x^t)$ for 
\begin{equation} \label{eq:alpha}
    \alpha_{\xi}(x) = \min \cbr{1, \frac{\tau(x)}{\norm{\nabla f_{\xi}(x)}}},
\end{equation}
and $\tau(x)$ is $p$-th quantile (of random variable $\norm{\gf(x)}$) clipping threshold, defined as 
\begin{equation} \label{eq:quantile}
    \mathrm{Prob}(\norm{\gf(x)} \leq \tau(x)) = p.
\end{equation}
We use the name SGD with Quantile Clipping (\alg) for the described algorithm. Note that Clipped \sgd is a special case of \alg for $\tau(x) \equiv \tau$.
This algorithm was originally introduced in the seminal work by \citet{merad2024robust} in the context of robust optimization with a corrupted oracle. They analyzed it as a Markov chain, while we are interested in optimization properties with a focus on differentially private settings.

\subsection{Preliminaries}

At first, we present some crucial properties of the described gradient estimator $g(x^t)$ and clipping threshold $\tau(x)$ needed for convergence analysis.

\begin{lemma}[\citet{merad2024robust}] \label{lemma:key_lemma}
Assume that stochastic gradient estimator $\gf(x)$ satisfies Assumption \ref{ass:b_variance}, $\alpha_{\xi}(x)$ is chosen as \eqref{eq:alpha}, and $p$-th quantile clipping threshold $\tau(x)$ satisfies \eqref{eq:quantile}. Then for all $x \in \RR^d$,
\begin{equation} \label{lemma:tau}
    \tau(x) \leq \norm{\nabla f(x)} + \sigma_q \br{1-p}^{-1/q},
\end{equation}
\begin{equation} \label{lemma:bias_}
    \norm{\Exp{\alpha_{\xi}(x) \gf(x)} - \oalpha(x) \nabla f(x)} \leq \sigma_q \br{1-p}^{1-1/q},
\end{equation}
where $\oalpha(x) \eqdef \Exp{\alpha_{\xi}(x) }$.
\end{lemma}
Note that \eqref{lemma:bias_} is different from typical results \cite{zhang2020adaptive, koloskova2023revisiting} characterizing the bias of gradient clipping due to the additional multiplier $\oalpha(x)$ before the gradient $\nabla f(x)$. This significantly impacts the convergence analysis of \alg and makes it different from Clipped \sgd.

\subsection{Convergence analysis}

Our analysis relies on the following recursion.

\begin{lemma} \label{lemma:recursion}
Suppose $f$ is $L$-smooth (\ref{ass:l_smooth}) and stochastic gradients satisfy Assumption \ref{ass:b_variance}.
Then, for $\beta>0$ the iterates of \alg \eqref{eq:sgd_qc} satisfy
\begin{align} \label{eq:core_recursion}
    \Exp{f(x^{t+1}) \mid x^t} &\leq f(x^t) - \gamma_t \br{\oalpha(x^t) - \beta/2 - \gamma_t L} \sqn{\nabla f(x^t)}
    \\&\qquad+ \frac{\gamma_t}{2} \beta^{-1}\br{1-p}^{2-2/q} \sigma_q^2 + \gamma^2_t L \sigma_q^2\br{1-p}^{-2/q} \notag
\end{align}
\end{lemma}
Equipped with Lemma \ref{lemma:recursion}, we provide a general convergence result for \alg.

\begin{theorem}[General case] \label{thm:non_cvx_gen}
Suppose $f$ is $L$-smooth (\ref{ass:l_smooth}) and stochastic gradients satisfy Assumption~\ref{ass:b_variance}. Then, for the step size chosen as
\begin{equation} \label{eq:step_size}
    0 < \gamma_t \leq \frac{2p - \beta - c}{2L},
\end{equation}
where $c, \beta \in (0, 1)$ and $\beta + c \leq 2p$ the iterates of \alg \eqref{eq:sgd_qc} satisfy
\begin{equation} \label{eq:gen_res}
    \frac{c}{\Gamma_T} \sum_{t=0}^{T-1} \gamma_t \Exp{\sqn{\nabla f(x^t)}}
    \leq 
    \frac{2\br{f(x^0) - f^{\inf}}}{\Gamma_T}
    + \frac{\sigma_q^2}{\Gamma_T} \sum_{t=0}^{T-1} \gamma_t h^{-2/q} \br{2 L\gamma_t + \beta^{-1}h^{2}},
\end{equation}
where $\Gamma_T \eqdef \sum_{t=0}^{T-1} \gamma_t$ and $h \eqdef 1 - p$.
\end{theorem}
Theorem \ref{thm:non_cvx_gen} indicates that \alg can find an approximate stationary point. Importantly, more aggressive clipping (smaller $p$) requires decreasing the step size $\gamma_t$ according to \eqref{eq:step_size}.
However, overall performance heavily depends on $h$ and the sequence $\gamma_t$. Next, we discuss how two possible choices affect convergence.

\begin{corollary}[Constant parameters] \label{cor:const_res}
For constant step size $\gamma_t \equiv \gamma \leq (2p - \beta - c)/(2L)$ convergence bound \eqref{eq:gen_res} reduces to
\begin{equation} \label{eq:constant_res}
    \frac{c}{T} \sum_{t=0}^{T-1} \Exp{\sqn{\nabla f(x^t)}} \leq 
    \underbrace{\frac{2\br{f(x^0) - f^{\inf}}}{\gamma T}}_{\circledOne}
    + \underbrace{2 \gamma L \sigma_q^2 h^{-2/q}}_{\circledTwo} + \underbrace{\beta^{-1} \sigma_q^2 h^{2-2/q}}_{\circledThree}.
\end{equation}
\end{corollary}
The obtained result shares similarities with classical \sgd convergence guarantees (under bounded stochastic gradient variance assumption) as it indicates convergence to the neighborhood of the stationary point. Namely, the first term \circledOne in upper bound \eqref{eq:constant_res} is basically the same and decreases with rate $\mathcal{O}(1/T)$. The second term \circledTwo is larger by factor $h^{-2/q}$ as $h = 1 - p\in [0, 1)$ indicating that more aggressive clipping (smaller $p$) increases the neighborhood.
However, the fundamental difference to \sgd is that exact convergence to the stationary point (making gradient arbitrary small: $\EE\sqn{\nabla f(x^t)} \leq \varepsilon^2$) cannot be guaranteed for any constant step size $\gamma$ as the third term \circledThree in \eqref{eq:constant_res} cannot be controlled. This means that the method has an irreducible bias, unlike standard \sgd, which enjoys convergence rate $\mathcal{O}(1/\sqrt{T})$ by choosing step size $\gamma$ as $\mathcal{O}(1/\sqrt{T})$.

In addition, our analysis indicates a trade-off between convergence speed and the size of the neighborhood. Specifically, larger $\beta$ decreases the latter (third term \circledThree in $\eqref{eq:constant_res}$) but requires choosing a smaller step size\footnote{And potentially a larger $p$ to ensure positive step size $\gamma > 0$.} $\gamma \leq \mathcal{O}\br{p-\beta}$ leading to slower convergence due to the first term \circledOne in \eqref{eq:constant_res} inversely proportional to $\gamma$.

\paragraph{Time-varying parameters.}
Using Theorem \ref{thm:non_cvx_gen}, we can jointly design a schedule of step size $\gamma_t$ and quantile values $p_t = 1 - h_t$ to guarantee convergence to the stationary point. 
Namely, for $\gamma_t = \tilde{\cO}(t^{\theta-1})$ and $h_t = \tilde{\cO}(t^{\nu})$ we have $\Gamma_T = \tilde{\cO}(T^\theta)$ and the upper bound \eqref{eq:gen_res} will be of order\footnote{We use $\tilde{\cO}$ notation to suppress constants other than $t, T$.}
\begin{equation}
    \tilde{\cO}\br{T^{-\theta} + T^{\theta-1-2\nu/q} + T^{2\nu(1-1/q)}},
\end{equation}
which is minimized for $\theta = (1-q^{-1})/(2-q^{-1})$ and $\nu = -1/(4-2q^{-1})$ resulting in complexity $\tilde{\cO}(\varepsilon^{-\frac{2(2q-1)}{q-1}})$. Thus, for standard bounded variance case $q=2$ the step size has to be decreased as $\gamma_t = \tilde{\cO}(t^{-2/3})$ and quantile increased as $p_t = 1 - \tilde{\cO}(t^{-1/3})$ to obtain convergence of order $\tilde{\cO}(T^{-1/3})$. This confirms the intuition that the method can converge right to the stationary point if clipping bias is eventually eliminated. However, our result does not necessarily require increasing the clipping threshold as norms of stochastic gradients $\norm{\gf(x^t)}$ may not converge to zero for increasing $t$.

\subsection{Comparison to fixed clipping}

The latest analysis on \sgd with constant clipping ($\tau(x) \equiv \tau$) we are aware of is due to \citet{koloskova2023revisiting}. Their (simplified) result indicates that with a proper step size choice, for $q=2, \sigma_q = \sigma$, and for $L$-smooth function, the expected squared gradient norm is upper bounded by
\begin{equation} \label{eq:const_clip}
\cO \br{\left(\frac{F^0}{\gamma T \tau}\right)^2 + \frac{F^0}{\gamma T} + \gamma L \sigma^2 + \min \br{\sigma^2, \frac{\sigma^4}{\tau^2}}},
\end{equation}
where $F^0 \eqdef f(x^0) - f^{\inf}$.
This result is fundamentally similar to quantile clipping \eqref{eq:constant_res} as the last term is also irreducible via decreasing step size $\gamma$.
However, upper bound \eqref{eq:const_clip} can be made arbitrary small by choosing step size as $\gamma = \tilde{\cO}(T^{-1/2})$ and clipping threshold as $\tau = \tilde{\cO}(T^\lambda), \lambda \in (0, 1/4)$. While increasing $\tau$ can solve the problem in theory, it is not satisfying from a practical perspective. As shown by \citet{andrew2021differentially}, the evolution of the distribution of the norm of the updates (or pseudo-gradients \cite{reddi2021adaptive}) may show very different behavior in federated training. Figure \ref{fig:quantile_clip} shows that norms of the updates may, in fact, increase during optimization.
In addition, for differentially private settings, bigger $\tau$ requires adding larger noise at every iteration, resulting in the degraded utility of the model \cite{bu2024automatic}.

\subsection{Bias due to clipping}
In the discussion after Theorem \ref{cor:const_res}, we mentioned that our result indicates that for any non-trivial fixed quantile $p \in (0, 1)$, exact convergence to the stationary point cannot be guaranteed for any step size $\gamma$. To demonstrate that this effect is not just a result of our (potentially suboptimal) analysis but that the method's estimator is indeed limited, we present the following function (based on \cite{koloskova2023revisiting}).
\begin{example}
For $r > 0$ and $\omega \in (1/2, 1)$ define
\begin{equation} \label{eq:example_f}
f_{\xi}(x)=\frac{1}{2}
\left\{\begin{array}{cc}
    (x+r)^2, &\text{with probability } \omega \\
    x^2, &\text{with probability } 1-\omega.
\end{array}\right.
\end{equation}
Then $\nabla f(x) = \Exp{\nabla f_{\xi}(x)} = x + r \omega$, which brings minima for $f(x)= \Exp{f_{\xi}(x)}$ at $x^\star = -r \omega$.
\end{example}
Suppose quantile $p$ is chosen so that half of the stochastic gradients are clipped at every point $x$ (e.g., as the median). Then, the estimator has the form
\begin{equation}
g(x) \eqdef \alpha_{\xi}(x) \nabla f_{\xi} (x) = 
\left\{\begin{array}{cc}
    1, &\text{with probability } \omega \\
    x, &\text{with probability } 1-\omega,
\end{array}\right.
\end{equation}
which indicates that $x^\dagger = -\omega/(1-\omega)$ is the expected fixed point of \alg as $\Exp{g(x^\dagger)} = 0$. Thus, if \alg converges, it must do so towards its fixed points. However, for any $r \neq 1/(1-\omega)$ minimum of $f$ is different from the expected fixed point $x^\star \neq x^\dagger$ and $\|\nabla f(x^\dagger)\| > 0$.

\section{Differentially Private Extension} \label{sec:dp_sgd}
The most standard way to make clipped \sgd $\br{\epsilon, \delta}$-Differentially Private (DP) is by adding isotropic Gaussian noise with variance proportional to the clipping threshold \cite{abadi2016deep} (along with subsampling/mini-batching). This approach applied to \alg results in the following update (\algname{DP-QC-SGD} for short)
\begin{equation} \label{eq:dp_sgd_qc}
    x^{t+1} = x^t - \gamma_t \frac{1}{B} \sum_{j=1}^B \br{g^t_j + z^t}, \qquad
    g^t_j = \min \biggl\{1, \frac{\tau(x^t)}{\bigl\|\nabla f_{\xi^t_j}(x^t)\bigr\|}\biggr\} \nabla f_{\xi^{t}_j}(x^t),
\end{equation}
where $z^t \sim \mathcal{N} (0, \br{\tau(x^t)}^2 \sigma^2_{\mathrm{DP}} \mathbf{I})$ and $\sigma_{\mathrm{DP}} \geq C \sqrt{T \log(\nicefrac{1}{\delta})} \epsilon^{-1}$ for some universal constant $C$ independent of $T, \delta, \epsilon$. For simplicity, we assume that $\xi_j^t$ are uniformly and independently sampled. 
In contrast to \algname{DP-SGD} with fixed clipping, the noise variance in our approach varies throughout training as it depends on the $\tau(x^t)$. This variability is crucial to the convergence result outlined in the upcoming theorem, setting our work apart from earlier studies \cite{koloskova2023revisiting}.

\begin{theorem}[\algname{DP-QC-SGD}] \label{thm:dp_sgd}
Suppose $f$ is $L$-smooth (\ref{ass:l_smooth}) and stochastic gradients satisfy Assumption~\ref{ass:b_variance}. Then, for the step size chosen as
\begin{equation} \label{eq:dp_step_size}
    \gamma_t \leq \frac{p - \beta/2 - c}{2 L {\color{blue}\mathfrak{S}}},
\end{equation}
where ${\color{blue}\mathfrak{S}} \eqdef \nicefrac{1}{B} + \sigma^2_{\mathrm{DP}}$, and $c, \beta \in (0, 1)$, and $\beta/2 + c \leq p$ the iterates of \algname{DP-QC-SGD} \eqref{eq:dp_sgd_qc} satisfy
\begin{equation} \label{eq:dp_sgd_res}
    \frac{c}{\Gamma_T} \sum_{t=0}^{T-1} \gamma_t \Exp{\sqn{\nabla f(x^t)}}
    \leq 
    \frac{f(x^0) - f^{\inf}}{\Gamma_T}
    + \frac{\sigma_q^2}{\Gamma_T} \sum_{t=0}^{T-1} \gamma_t h^{-2/q} \br{2 \gamma_t L {\color{blue}\mathfrak{S}} + \beta^{-1}h^{2}/2},
\end{equation}
where $\Gamma_T \eqdef \sum_{t=0}^{T-1} \gamma_t$ and $h \eqdef 1 - p$.
\end{theorem}
Analogously to Corollary \ref{cor:const_res}, we simplify the result for constant step size $\gamma_t \equiv \gamma$.
\begin{equation} \label{eq:const_dp_sgd_res}
    \frac{c}{T} \sum_{t=0}^{T-1} \Exp{\sqn{\nabla f(x^t)}}
    \leq 
    \frac{f(x^0) - f^{\inf}}{\gamma T}
    + 2 \gamma L \sigma_q^2 h^{-2/q} {\color{blue}\mathfrak{S}} 
    + \beta^{-1} \sigma_q^2 h^{2-2/q}/2.
\end{equation}
Theorem \ref{thm:dp_sgd} is similar to non-private result \eqref{eq:gen_res} in nature as it shows convergence to a neighborhood of the stationary point. However, there is a significant difference expressed in term ${\color{blue}\mathfrak{S}} = \nicefrac{1}{B} + \sigma^2_{\mathrm{DP}}$ in the denominator of the step size condition \eqref{eq:dp_step_size} and convergence bound \eqref{eq:const_dp_sgd_res}.
Namely, mini-batching decreases the stochastic term \circledTwo but does not affect the \say{bias} part \circledThree in \eqref{eq:constant_res}.
Note that ${\color{blue}\mathfrak{S}}$ can be even smaller than 1 in private federated learning for a big enough cohort size $B$ and a small number of communication rounds $T$. However, for centralized DP training, ${\color{blue}\mathfrak{S}}$ is likely to be larger, which results in a smaller step size and larger convergence neighborhood. The latter, though, can be eliminated via a standard \sgd step size strategy as the term in \eqref{eq:const_dp_sgd_res} involving ${\color{blue}\mathfrak{S}}$ depends on $\gamma$.

\section{Conclusion and Future Directions}
Our work provides the first theoretical analysis of \sgd with quantile clipping, advancing the understanding of adaptive clipping methods. We established convergence guarantees for \alg under standard smoothness and heavy-tailed noise conditions.
Our analysis revealed that \alg suffers from an inherent bias problem analogous to that observed in fixed clipping \sgd. Furthermore, we designed a time-varying quantile and step-size schedule that could fix the discovered limitation in a practical way.
Finally, a differentially private extension: \algname{DP-QC-SGD} method was proposed and analyzed.

Moving forward, the observed limitations of the \alg (with fixed quantile) raise the question of possible improvements via algorithmic modifications.
It is also worth noting that the current analysis is performed for an idealized case when the exact quantile $\tau(x)$ is available. This may not be feasible in certain practical scenarios, such as cross-device FL, that typically allow access to an approximation.
Moreover, despite the empirical success of adaptive clipping, there are scenarios where it performs suboptimally \cite{xu2023federated}, motivating future research.
Addressing these open questions can lead to more robust and efficient private learning methods.

\section*{Acknowledgments and Disclosure of Funding}
We would like to thank anonymous reviewers for their valuable comments, which improved the manuscript, and Abdurakhmon Sadiev for helpful technical discussions.

The research reported in this publication was supported by funding from King Abdullah University of Science and Technology (KAUST): i) KAUST Baseline Research Scheme, ii) Center of Excellence for Generative AI, under award number 5940, iii) SDAIA-KAUST Center of Excellence in Artificial Intelligence and Data Science.

\bibliographystyle{abbrvnat}
\bibliography{refs}

\appendix

\part*{Appendix}

\tableofcontents

\section{Basic and Auxiliary Facts}

For all vectors $a, b\in \Rd$ and $\beta > 0$: 
\begin{equation} \label{eq:triangle}
    \norm{a + b} \leq \norm{a} + \norm{b},
\end{equation}
\begin{equation} \label{eq:triangle_square}
    \sqn{a + b} \leq 2\sqn{a} + 2\sqn{b},
\end{equation}
\begin{equation} \label{eq:cauchy–schwarz}
    |\<a, b>| \leq \norm{a} \norm{b},
\end{equation}
\begin{equation} \label{eq:fenchel_young}
    2\<a, b> \leq \beta \sqn{a} + \beta^{-1} \sqn{b}.
\end{equation}

\section{Proofs}
\subsection{Proof of Lemma \ref{lemma:key_lemma}}
We provide the proof adapted from \cite[Lemma 2]{merad2024robust} for completeness.
Note that 
\begin{equation}
    \II\cbr{\norm{\gf(x)} \leq \tau(x)} + \II\cbr{\norm{\gf(x)} > \tau(x)} = 1.
\end{equation}
\textbf{1.} As a reminder $\oalpha(x) \eqdef \Exp{\alpha_{\xi}(x)}$. Then
\begin{align*}
    \Exp{\alpha_{\xi}(x) \gf(x)} - \oalpha(x) \nabla f(x) 
    &=
    \Exp{\br{\alpha_{\xi}(x) - \oalpha(x)} \br{\gf(x) - \nabla f(x)}}
    \\&=
    \Exp{\br{\alpha_{\xi}(x) - \oalpha(x)} \br{\gf(x) - \nabla f(x)} \II\cbr{\norm{\gf(x)} \leq \tau(x)}} 
    \\&\qquad + 
    \Exp{\br{\alpha_{\xi}(x) - \oalpha(x)} \br{\gf(x) - \nabla f(x)} \II\cbr{\norm{\gf(x)} > \tau(x)}}
    \\&\overset{\eqref{eq:alpha}}{=}
    \Exp{\br{1 - \oalpha(x)} \br{\gf(x) - \nabla f(x)} \br{1 - \II\cbr{\norm{\gf(x)} > \tau(x)}}} 
    \\&\qquad - 
    \oalpha(x)\Exp{ \br{\gf(x) - \nabla f(x)} \II\cbr{\norm{\gf(x)} > \tau(x)}}
    \\&\qquad + 
    \Exp{\frac{\tau(x)}{\norm{\gf(x)}} \br{\gf(x) - \nabla f(x)} \II\cbr{\norm{\gf(x)} > \tau(x)}}
    \\&=
    \Exp{\br{1 - \oalpha(x)} \br{\gf(x) - \nabla f(x)}} 
    \\&\qquad + 
    \Exp{\br{\oalpha(x) - 1} \br{\gf(x) - \nabla f(x)} \II\cbr{\norm{\gf(x)} > \tau(x)}}
    \\&\qquad - 
    \oalpha(x)\Exp{ \br{\gf(x) - \nabla f(x)} \II\cbr{\norm{\gf(x)} > \tau(x)}}
    \\&\qquad + 
    \Exp{ \frac{\tau(x)}{\norm{\gf(x)}} \br{\gf(x) - \nabla f(x)} \II\cbr{\norm{\gf(x)} > \tau(x)}}
    \\&=
    \Exp{\br{\frac{\tau(x)}{\norm{\gf(x)}} - 1} \br{\gf(x) - \nabla f(x)} \II\cbr{\norm{\gf(x)} > \tau(x)}}.
\end{align*}
Next, we use the fact that if $\frac{\tau(x)}{\norm{\gf(x)}} \in (0, 1)$ then $|\tau(x)/\norm{\gf(x)} - 1| \in (0, 1)$. Thus,
\begin{align*}
    \norm{\Exp{\alpha_{\xi}(x) \gf(x)} - \oalpha(x) \nabla f(x)}
    &\leq
    \Exp{\left|\frac{\tau(x)}{\norm{\gf(x)}} - 1\right| \II\cbr{\norm{\gf(x)} > \tau(x)} \norm{\gf(x) - \nabla f(x)}}
    \\&\leq
    \Exp{\II\cbr{\norm{\gf(x)} > \tau(x)} \norm{\gf(x) - \nabla f(x)}}
    \\&\overset{(*)}{\leq}
    \br{1-p}^{1-1/q} \Exp{\norm{\gf(x) - \nabla f(x)}^q}^{1/q}
    \\&\overset{\eqref{eq:bound_var}}{\leq}
    \br{1-p}^{1-1/q} \sigma_q,
\end{align*}
where $(*)$ is due to Hölder's inequality.

\textbf{2.} Denote by $\cQ_p\br{\norm{\gf(x)}} = \tau(x)$ the $p$-th quantile of $\norm{\gf(x)}$ distribution. Then
\begin{equation}
    \tau(x) = 
    \cQ_p\br{\norm{\gf(x) - \nabla f(x) + \nabla f(x)}} \overset{\eqref{eq:triangle}}{\leq}
    \norm{\nabla f(x)} + \cQ_p\Bigl(\bigl\|\underbrace{\gf(x) - \nabla f(x)}_{\delta_x}\bigr\|\Bigr).
\end{equation}
By using quantile definition and Markov's inequality
\begin{equation}
    1 - p = 
    \mathrm{Prob} \cbr{\norm{\delta_x} > \cQ_p(\norm{\delta_x})} \leq
    \br{\frac{\Exp{\norm{\delta_x}}}{\cQ_p(\norm{\delta_x})}}^q \leq
    \br{\frac{\sigma_q}{\cQ_p(\norm{\delta_x})}}^q,
\end{equation}
where the last inequality holds for $q>1$ as $\br{\Exp{\norm{\delta_x}}}^q \leq \Exp{\norm{\delta_x}^q}$.
Therefore $\cQ_p(\norm{\delta_x}) \leq \frac{\sigma_q}{\br{1-p}^{1/q}}$.

\subsection{Proof of Lemma \ref{lemma:recursion}} \label{sec:rec_lemma_proof}
By using $L$-smoothness (\ref{ass:l_smooth}) for iterates of algorithm \eqref{eq:sgd_qc}
\begin{equation} \label{eq:sgd_qc_full}
    x^{t+1} = x^t - \gamma_t g^t, \qquad g^t = \alpha_{\xi^t}(x^t) \nabla f_{\xi}(x^t) =  \min \cbr{1, \frac{\tau(x^t)}{\norm{\nabla f_{\xi}({x^t})}}} \nabla f_{\xi}(x^t).
\end{equation}
we have\footnote{In the proofs, we omit expectation condition in some cases for brevity.}
\begin{align} \label{eq:descent_lemma_1} 
    \Exp{f(x^{t+1}) \mid x^t} 
    &\leq f(x^t)
    - \gamma_t \inner{\nabla f(x^t)}{\Exp{g^t}} + \frac{\gamma^2_t L}{2} \Exp{\sqn{g^t}}
    \\ &\leq \notag
    f(x^t) - \gamma_t \inner{\nabla f(x^t)}{\Exp{g^t} \pm \oalpha(x^t) \nabla f(x^t)} + \frac{\gamma^2_t L}{2} \Exp{\sqn{g^t}}
    \\ &\leq \notag
    f(x^t) - \gamma_t \oalpha(x^t) \sqn{\nabla f(x^t)}
    - \gamma_t \inner{\nabla f(x^t)}{\Exp{g^t} - \oalpha(x^t) \nabla f(x^t)} + \frac{\gamma^2_t L}{2} \Exp{\sqn{g^t}}
    \\ &\overset{\eqref{eq:cauchy–schwarz}}{\leq} \notag
    f(x^t) - \gamma_t \oalpha(x^t) \sqn{\nabla f(x^t)}
    + \gamma_t \norm{\nabla f(x^t)}\underbrace{\norm{\Exp{g^t} - \oalpha(x^t) \nabla f(x^t)}}_{G_t} + \frac{\gamma^2_t L}{2} \Exp{\sqn{g^t}}
    \\ &\overset{\eqref{eq:fenchel_young}}{\leq} \notag
    f(x^t) - \gamma_t \oalpha(x^t) \sqn{\nabla f(x^t)}
    + \frac{\gamma_t}{2} \br{\beta\sqn{\nabla f(x^t)} + \beta^{-1} G_t^2} + \frac{\gamma^2_t L}{2} \Exp{\sqn{g^t}}
    \\ &\overset{\eqref{lemma:bias_}}{\leq} \notag
    f(x^t) - \gamma_t \br{\oalpha(x^t) - \beta/2} \sqn{\nabla f(x^t)}
    + \frac{\gamma_t}{2}\beta^{-1} \sigma_q^2\br{1-p}^{2-2/q} + \frac{\gamma^2_t L}{2} \Exp{\sqn{g^t}},
\end{align}
where in the last but one step, we used Fenchel-Young inequality for $\beta > 0$. Next, we upper bound the second moment $\EE\sqn{g^t}$.
\begin{align*}
    \Exp{f(x^{t+1}) \mid x^t} 
    &\leq
    f(x^t) - \gamma_t \br{\oalpha(x^t) - \beta/2} \sqn{\nabla f(x^t)}
    + \frac{\gamma_t}{2}\beta^{-1} \sigma_q^2\br{1-p}^{2-2/q} + \frac{\gamma^2_t L}{2} \Exp{\sqn{g^t}}
    \\ &\overset{\eqref{eq:sgd_qc_full}}{\leq}
    f(x^t) - \gamma_t \br{\oalpha(x^t) - \beta/2} \sqn{\nabla f(x^t)}
    + \frac{\gamma_t}{2}\beta^{-1} \sigma_q^2\br{1-p}^{2-2/q} + \frac{\gamma^2_t L}{2} \br{\tau(x^t)}^2
    \\ &\overset{\eqref{lemma:tau}}{\leq}
    f(x^t) - \gamma_t \br{\oalpha(x^t) - \beta/2} \sqn{\nabla f(x^t)}
    + \frac{\gamma_t}{2}\beta^{-1} \sigma_q^2\br{1-p}^{2-2/q} 
    \\&\qquad + \frac{\gamma^2_t L}{2} \br{\norm{\nabla f(x^t)} + \sigma_q\br{1-p}^{-1/q}}^2
    \\ &\overset{\eqref{eq:triangle_square}}{\leq}
    f(x^t) - \gamma_t \br{\oalpha(x^t) - \gamma_t L - \beta/2} \sqn{\nabla f(x^t)}
    + \frac{\gamma_t}{2}\beta^{-1} \sigma_q^2\br{1-p}^{2-2/q} + \gamma^2_t L \sigma_q^2\br{1-p}^{-2/q}.
\end{align*}
Rearranging the terms yields the desired result.

\subsection{Proof of Theorem \ref{thm:non_cvx_gen} (\alg)} \label{appendix:qc_sgd}
Denote $h \eqdef 1 - p$, then Lemma \ref{lemma:recursion} (with suppressed expectation condition) gives
\begin{align*}
    \Exp{f(x^{t+1})} - f(x^t) 
    &\leq - \gamma_t \br{\oalpha(x^t) - \beta/2 - \gamma_t L} \sqn{\nabla f(x^t)}
    + \frac{\gamma_t}{2} \beta^{-1}h^{2-2/q} \sigma_q^2 + \gamma^2_t L \sigma_q^2 h^{-2/q}
    \\ &\leq
    - \gamma_t \br{p - \beta/2 - \gamma_t L} \sqn{\nabla f(x^t)}
    + \frac{\gamma_t}{2} \beta^{-1}h^{2-2/q} \sigma_q^2 + \gamma^2_t L \sigma_q^2 h^{-2/q}
\end{align*}
where we also used the fact that $\oalpha(x) \geq p$. Next, we choose the step size as 
\begin{equation*}
    0 \leq \gamma_t \leq \frac{2p - \beta - c}{2L}
\end{equation*}
to enforce condition $p - \beta/2 - \gamma_t L \geq c/2$. This leads to
\begin{equation*}
    \Exp{f(x^{t+1})} - f(x^t) \leq - \frac{c}{2} \gamma_t \sqn{\nabla f(x^t)}
    + \frac{\gamma_t}{2} \beta^{-1}h^{2-2/q} \sigma_q^2 + \gamma^2_t L \sigma_q^2h^{-2/q}.
\end{equation*}
After rearranging the terms, we have a recursion
\begin{equation*}
    c\gamma_t \sqn{\nabla f(x^t)} \leq 2\br{f(x^t) - \Exp{f(x^{t+1})}} 
    + \sigma_q^2 \gamma_t h^{-2/q} \br{\beta^{-1} h^2 + 2 \gamma_t L}.
\end{equation*}
Summing over $t$ from $0$ to $T-1$ and unrolling the recursion leads to 
\begin{align*}
    c \sum_{t=0}^{T-1} \gamma_t \Exp{\sqn{\nabla f(x^t)}}
    &\leq 
    \sum_{t=0}^{T-1} 2\br{f(x^0) - \Exp{f(x^{t+1})}}
    + \sigma_q^2 \sum_{t=0}^{T-1} \gamma_t h^{-2/q} \br{2 L\gamma_t + \beta^{-1}h^{2}}
    \\&\leq
    2\br{f(x^0) - \Exp{f(x^{T})}}
    + \sigma_q^2 \sum_{t=0}^{T-1} \gamma_t h^{-2/q} \br{2 L\gamma_t + \beta^{-1}h^{2}}
    \\&\overset{\eqref{ass:l_smooth}}{\leq}
    2\br{f(x^0) - f^{\inf}}
    + \sigma_q^2 \sum_{t=0}^{T-1} \gamma_t h^{-2/q} \br{2 L\gamma_t + \beta^{-1}h^{2}}.
\end{align*}
Dividing over $\Gamma_T = \sum_{t=0}^{T-1} \gamma_t$ yields the final result
\begin{equation*}
    \frac{c}{\Gamma_T} \sum_{t=0}^{T-1} \gamma_t \Exp{\sqn{\nabla f(x^t)}}
    \leq 
    \frac{2\br{f(x^0) - f^{\inf}}}{\Gamma_T}
    + \frac{\sigma_q^2}{\Gamma_T} \sum_{t=0}^{T-1} \gamma_t h^{-2/q} \br{2 L\gamma_t + \beta^{-1}h^{2}}.
\end{equation*}

\subsection{Proof of Theorem \ref{thm:dp_sgd} (\algname{DP-QC-SGD})} \label{appendix:dp_sgd}
As a reminder, the original method \eqref{eq:sgd_qc} is changed via mini-batching and adding Gaussian noise.

\begin{equation} \label{eq:dp_ssgd_qc_appndx}
    x^{t+1} = x^t - \gamma_t \underbrace{\frac{1}{B} \sum_{j=1}^B \br{g^t_j + z^t}}_{\tilde{g}^t}, \qquad
    g^t_j = \min \biggl\{1, \frac{\tau(x^t)}{\bigl\|\nabla f_{\xi^t_j}(x^t)\bigr\|}\biggr\} \nabla f_{\xi^{t}_j}(x^t),
\end{equation}
where $z^t \sim \mathcal{N} \br{0, \br{\tau(x^t)}^2 \sigma^2_{\mathrm{DP}} \mathbf{I}}$ are sampled i.i.d.
\begin{proof}
By inspecting the proof of Lemma \ref{sec:rec_lemma_proof}, namely $L$-smoothness inequality
\begin{equation} \label{eq:dp_descent}
    \Exp{f(x^{t+1}) \mid x^t} 
    \leq f(x^t)
    - \gamma_t \inner{\nabla f(x^t)}{\Exp{\tilde{g}^t} \pm \oalpha(x^t) \nabla f(x^t)} + \frac{\gamma^2_t L}{2} \Exp{\sqn{\tilde{g}^t}}
\end{equation}
it is clear that the \algname{DP-SGD} extension impacts the last two terms with $\tilde{g}^t$ while for $g_j^t$ one can reuse the same results as for \alg. Next, we show how DP modification affects the second moment and ``bias'' of the gradient estimator.

1. Due to the independence of $\xi^t_j$, the second moment of the stochastic gradient estimator can be upper-bounded as
\begin{align} \label{eq:dp_sgd_1}
    \EE\sqn{\tilde{g}^t}
    &= 
    \EE\sqn{\frac{1}{B} \sum_{j=1}^B \br{g^t_j + z^t}}
    \overset{\eqref{eq:triangle_square}}{\leq}
    2\EE\sqn{\frac{1}{B} \sum_{j=1}^B g^t_j} + 2\EE\sqn{z^t} \leq
    \frac{2}{B^2} \sum_{j=1}^B \EE\sqn{g^t_j} + 2\EE\sqn{z^t} \notag
    \\&\overset{\eqref{eq:dp_ssgd_qc_appndx}}{\leq}
    \frac{2}{B^2} \sum_{j=1}^B \br{\tau(x^t)}^2 + 2 \br{\tau(x^t)}^2 \sigma^2_{\mathrm{DP}} =
    2 \br{\tau(x^t)}^2 \br{1/B + \sigma^2_{\mathrm{DP}}}.
\end{align}
2. Inequality \eqref{lemma:bias_} from Lemma \ref{lemma:key_lemma} is modified in the following way due to $\Exp{z^t} = 0$ for every $t$:
\begin{align} \label{eq:dp_sgd_2} \notag
    \norm{\Exp{\tilde{g}^t} - \oalpha(x^t) \nabla f(x^t)} &= 
    \norm{\frac{1}{B} \sum_{j=1}^B \Exp{g^t_j} - \oalpha(x^t) \nabla f(x^t)} 
    \\&\overset{\eqref{eq:triangle}}{\leq} \notag
    \frac{1}{B} \sum_{j=1}^B \norm{\Exp{g^t_j} - \oalpha(x^t) \nabla f(x^t)} 
    \\&\overset{\eqref{lemma:bias_}}{\leq}
    \br{1-p}^{1-1/q} \sigma_q.
\end{align}
Thus, only the last (second moment) term in upper bound \eqref{eq:dp_descent} is affected.

Convergence proof repeats the first steps from Appendix \ref{sec:rec_lemma_proof} starting with inequality \eqref{eq:descent_lemma_1}
and is changed in the following way
\begin{align*}
    \Exp{f(x^{t+1}) \mid x^t} 
    &\leq 
    f(x^t) - \gamma_t \br{\oalpha(x^t) - \beta/2} \sqn{\nabla f(x^t)}
    + \frac{\gamma_t}{2}\beta^{-1} \sigma_q^2\br{1-p}^{2-2/q} + \frac{\gamma^2_t L}{2} \Exp{\sqn{\tilde{g}^t}}
    \\ &\overset{\eqref{eq:dp_sgd_1}}{\leq}
    f(x^t) - \gamma_t \br{\oalpha(x^t) - \beta/2} \sqn{\nabla f(x^t)} + \frac{\gamma_t}{2} \beta^{-1} \sigma_q^2 \br{1-p}^{2-2/q}
    \\& \qquad + \gamma^2_t L \br{1/B + \sigma^2_{\mathrm{DP}}} \br{\tau(x^t)}^2
    \\ &\overset{\eqref{lemma:tau}}{\leq}
    f(x^t) - \gamma_t \br{\oalpha(x^t) - \beta/2} \sqn{\nabla f(x^t)}
    + \frac{\gamma_t}{2} \beta^{-1} \sigma_q^2 \br{1-p}^{2-2/q}
    \\& \qquad + \gamma^2_t L \br{1/B + \sigma^2_{\mathrm{DP}}} \br{\norm{\nabla f(x^t)} + \sigma_q\br{1-p}^{-1/q}}^2
    \\ &\overset{\eqref{eq:triangle_square}}{\leq}
    f(x^t) - \gamma_t \br{\oalpha(x^t) - \beta/2 - 2\gamma_t L \br{1/B + \sigma^2_{\mathrm{DP}}}} \sqn{\nabla f(x^t)}
    \\& \qquad + \frac{\gamma_t}{2} \beta^{-1} \sigma_q^2 \br{1-p}^{2-2/q}
    + 2 \gamma^2_t L \br{1/B + \sigma^2_{\mathrm{DP}}} \sigma_q^2 \br{1-p}^{-2/q}
    \\ &\leq
    f(x^t) - \gamma_t \br{p - \beta/2 - 2\gamma_t L \br{1/B + \sigma^2_{\mathrm{DP}}}} \sqn{\nabla f(x^t)}
    \\& \qquad + \frac{\gamma_t}{2} \beta^{-1} \sigma_q^2 \br{1-p}^{2-2/q}
    + 2 \gamma^2_t L \br{1/B + \sigma^2_{\mathrm{DP}}} \sigma_q^2 \br{1-p}^{-2/q},
\end{align*}
where the last inequality is due to the fact that $\oalpha(x) \geq p$.

Denote $\mathfrak{S} \eqdef 1/B+\sigma^2_{\mathrm{DP}}$, then modified step size condition would be
\begin{equation}
    \gamma_t \leq \frac{p - \beta/2 - c}{2 \mathfrak{S} L}.
\end{equation}
to guarantee that $p - \beta/2 - 2\gamma_t L \mathfrak{S} \geq c$. After denoting $h \eqdef 1-p$ we obtain
\begin{equation*}
    c \gamma_t \sqn{\nabla f(x^t)} \leq f(x^t) - \Exp{f(x^{t+1}) \mid x^t}
    + \frac{\gamma_t}{2} \beta^{-1} \sigma_q^2 h^{2-2/q}
    + 2 \gamma^2_t L \mathfrak{S} \sigma_q^2 h^{-2/q}.
\end{equation*}
Repeating the last steps from Appendix \ref{appendix:qc_sgd}, i.e. summing over $t$ from $0$ to $T-1$, unrolling the recursion, and dividing over $\Gamma_T = \sum_{t=0}^{T-1} \gamma_t$ leads to the final result
\begin{equation*}
    \frac{c}{\Gamma_T} \sum_{t=0}^{T-1} \gamma_t \Exp{\sqn{\nabla f(x^t)}}
    \leq 
    \frac{f(x^0) - f^{\inf}}{\Gamma_T}
    + \frac{\sigma_q^2}{\Gamma_T} \sum_{t=0}^{T-1} \gamma_t h^{-2/q} \br{\beta^{-1}h^{2}/2 + 2 \gamma_t L \mathfrak{S}}.
\end{equation*}
\end{proof}

\end{document}